\documentclass[10pt,twocolumn,letterpaper]{article}

\usepackage{cvpr}
\usepackage{times}
\usepackage{epsfig}
\usepackage{graphicx}
\usepackage{amsmath}
\usepackage{amssymb}

\usepackage{epsfig}
\usepackage{subfigure}
\usepackage{graphicx}
\usepackage{algorithm}
\usepackage{algorithmic}
\usepackage{amssymb}
\usepackage{mathrsfs} 
\usepackage{amsmath,bm}
\usepackage{array,multirow}
\usepackage{mdwlist}
\usepackage{paralist}
\usepackage{url}
\usepackage{color}
\usepackage{caption}
\usepackage{setspace}
\usepackage{textcomp}
\usepackage{bbding}
\usepackage{booktabs}
\usepackage{threeparttable}

\usepackage[pagebackref=true,breaklinks=true,letterpaper=true,colorlinks,bookmarks=false]{hyperref}

\cvprfinalcopy 

\def\cvprPaperID{1258} 

\ifcvprfinal\fi
\begin{document}

\title{Spatiotemporal CNN for Video Object Segmentation}

\author{Kai Xu$^{1}$, Longyin Wen$^{2}$, Guorong Li$^{1,3}$\thanks{Corresponding author.}, Liefeng Bo$^{2}$, Qingming Huang$^{1,3,4}$ \\
$^1$ School of Computer Science and Technology, UCAS, Beijing, China. \\
$^2$ JD Digits, Mountain View, CA, USA. \\
$^3$ Key Laboratory of Big Data Mining and Knowledge Management, CAS, Beijing, China. \\
$^4$ Key Laboratory of Intell. Info. Process. (IIP), Inst. of Computi. Tech., CAS, China. \\
{\tt\small xukai16@mails.ucas.ac.cn}, {\tt \small \{longyin.wen,liefeng.bo\}@jd.com}, {\tt \small \{qmhuang,liguorong\}@ucas.ac.cn}
}

\maketitle
\thispagestyle{empty}

\begin{abstract}
In this paper, we present a unified, end-to-end trainable spatiotemporal CNN model for VOS, which consists of two branches, \ie, the temporal coherence branch and the spatial segmentation branch. Specifically, the temporal coherence branch pretrained in an adversarial fashion from unlabeled video data, is designed to capture the dynamic appearance and motion cues of video sequences to guide object segmentation. The spatial segmentation branch focuses on segmenting objects accurately based on the learned appearance and motion cues. To obtain accurate segmentation results, we design a coarse-to-fine process to sequentially apply a designed attention module on multi-scale feature maps, and concatenate them to produce the final prediction. In this way, the spatial segmentation branch is enforced to gradually concentrate on object regions. These two branches are jointly fine-tuned on video segmentation sequences in an end-to-end manner. Several experiments are carried out on three challenging datasets (\ie, DAVIS-2016, DAVIS-2017 and Youtube-Object) to show that our method achieves favorable performance against the state-of-the-arts. Code is available at \url{https://github.com/longyin880815/STCNN}.
\end{abstract}

\section{Introduction}
Video object segmentation (VOS) becomes a hot topic in recent years, which is a crucial step for many video analysis tasks, such as video summarization, video editing, and scene understanding. It aims to extract foreground objects from video clips. Existing VOS methods can be divided into two settings based on the degrees of human involvement, namely, {\it unsupervised} and {\it semi-supervised}. The unsupervised VOS methods \cite{DBLP:conf/cvpr/XiaoFY16,DBLP:conf/cvpr/TokmakovAS17,DBLP:conf/eccv/HuHS18,DBLP:conf/eccv/LiSVLK18,DBLP:conf/eccv/KohLK18} do not require any manual annotation, while the semi-supervised methods \cite{DBLP:journals/corr/VoigtlaenderL17,DBLP:conf/cvpr/Cheng18,DBLP:conf/eccv/CiWW18,DBLP:conf/eccv/HuHS18a} rely on the annotated mask for objects in the first frame. In this paper, we are interested in the semi-supervised VOS task, which can be treated as the label propagation problem through the entire video. To maintain the temporal associations of object segments, optical flow is usually used in most of previous methods \cite{DBLP:conf/cvpr/WenDLLY15,DBLP:conf/cvpr/Tsai0B16,DBLP:conf/iccv/ChengTW017,DBLP:conf/cvpr/JangK17,DBLP:conf/cvpr/TokmakovAS17,DBLP:conf/cvpr/abs-1803-09453,DBLP:conf/cvpr/HuWKKT18} to model the pixel consistency across the time for smoothness. However, optical flow annotation requires significant human effort, and estimation is challenging and often inaccurate, and thus it is not always helpful in video segmentation. To that end, Li \etal \cite{DBLP:conf/eccv/LiL18} design an end-to-end trained deep recurrent network to segment and track objects in video simultaneously. Xu \etal \cite{DBLP:conf/eccv/XuYFYYLPCH18} present a sequence-to-sequence network to fully exploit long-term spatial-temporal information for VOS.

In contrast to the aforementioned methods, we design a spatiotemporal convolutional neural network (CNN) algorithm (denoted as STCNN, for short) for VOS, which is a unified, end-to-end trainable CNN. STCNN is formed by two branches, \ie, the temporal coherence branch and the spatial segmentation branch. The features in both branches are able to obtain useful gradient information during back-propagation. Specifically, the temporal coherence branch focuses on capturing the dynamic appearance and motion cues to provide the guidance of object segmentation, which is pre-trained in an adversarial manner from unlabeled video data following \cite{DBLP:conf/iccv/JinLXSLYCDLJFY17}. The spatial segmentation branch is a fully convolutional network focusing on segmenting objects based on the learned appearance and motion cues from the temporal coherence branch. Inspired by \cite{DBLP:conf/cvpr/HuWKKT18}, we design a coarse-to-fine process to sequentially apply a designed attention module on multi-scale feature maps, and concatenate them to produce the final accurate prediction. In this way, the spatial segmentation branch is enforced to gradually concentrate on the object regions, which benefits both training and testing. These two branches are jointly fine-tuned on the video segmentation sequences (\eg, the {\tt training} set in DAVIS-2016 \cite{DBLP:conf/cvpr/PerazziPMGGS16}) in an end-to-end manner. We conduct several experiments on three challenging datasets, \ie, DAVIS-2016 \cite{DBLP:conf/cvpr/PerazziPMGGS16}, DAVIS-2017 \cite{DBLP:journals/corr/Pont-TusetPCASG17} and Youtube-Object \cite{DBLP:conf/cvpr/PrestLCSF12,DBLP:conf/eccv/JainG14}, to demonstrate the effectiveness of the proposed method against the state-of-the-art methods. Specifically, our STCNN method produces $0.838$ in mIOU for semi-supervised task on the DAVIS-2016 \cite{DBLP:conf/cvpr/PerazziPMGGS16}, and achieves the state-of-the-art results with $0.796$ in mIoU on Youtube-Object \cite{DBLP:conf/cvpr/PrestLCSF12,DBLP:conf/eccv/JainG14}.

{\bf Contributions}. (1) We present a unified, end-to-end trainable spatiotemporal CNN algorithm for VOS without relying on optical flow, which is formed by two branches, \ie, spatial segmentation branch and temporal coherence branch. (2) The temporal coherence branch is designed to capture the dynamic appearance and motion cues across the time to guide object segmentation, which is pre-trained in an adversarial manner from unlabeled video data. (3) We design a coarse-to-fine process to sequentially apply a designed attention module on multi-scale features maps, and concatenate them to produce the final accurate prediction. (4) Extensive experiments are conducted on three datasets, namely, DAVIS-2016, DAVIS-2017, and Youtube-Object, to demonstrate that the proposed method achieves favorable performance compared to the state-of-the-arts.

\begin{figure*}
\centering
\includegraphics[width=0.98\linewidth]{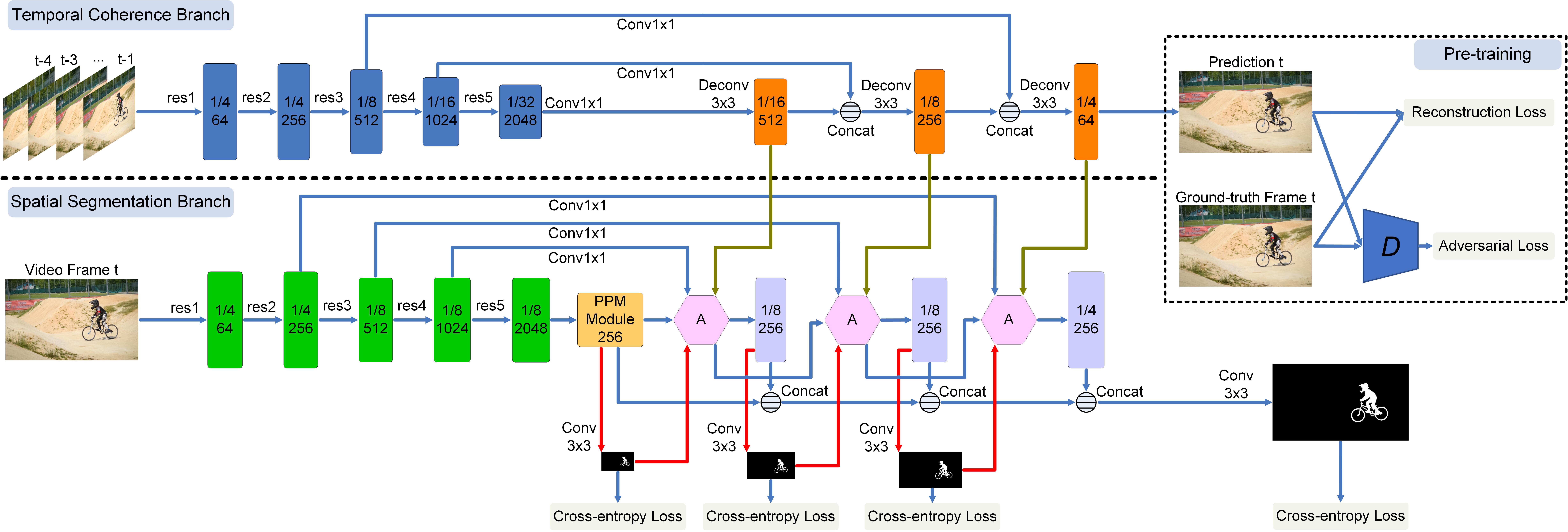}
\vspace{-2mm}
\caption{Overview of the network architecture of our STCNN algorithm. The part above the dashed line is the temporal coherence branch, and the part below the dashed line is the spatial segmentation branch. The red lines indicate the attention mechanism used in our model, and the hexagon indicates the attention module. Notably, each convolution layer is followed by a batch normalization layer \cite{DBLP:conf/icml/IoffeS15} and a ReLU layer.}
\label{fig:architecture}
\end{figure*}

\section{Related Work}
{\noindent \textbf{Semi-supervised video object segmentation.}}
Semi-supervised VOS aims to segment video objects based on the preliminarily provided foreground regions, and propagates them to the remaining frames. In \cite{DBLP:journals/pami/BadrinarayananBC13}, a patch-based probabilistic graphical model is presented for semi-supervised VOS, which uses a temporal tree structure to link patches in adjacent frames to exactly infer the pixel labels in video. Jain \etal \cite{DBLP:conf/eccv/JainG14} design a higher-order supervoxel label consistency potential for foreground region propagation, which leverages bottom-up supervoxels to guide the estimation towards long-range coherent regions. Wen \etal \cite{DBLP:conf/cvpr/WenDLLY15} integrate the multi-part tracking and segmentation into a unified energy objective to handle the VOS, which is efficiently solved by a RANSAC-style approach. Tsai \etal \cite{DBLP:conf/cvpr/Tsai0B16} jointly optimize VOS and optical flow estimation in a unified framework using an iterative scheme to exploit mutually bootstrapping information between the two tasks for better performance.

Recently, the deep neural network based methods dominate the VOS task. Khoreva \etal \cite{DBLP:conf/cvpr/Khoreva17} describe a CNN-based algorithm, which combines offline and online learning strategies, where the former produces a refined mask from the estimation of previous frame, and the latter aims to capture the appearance of the specific object instance. Cheng \etal \cite{DBLP:conf/iccv/ChengTW017} presents an end-to-end trainable network for simultaneously predicting pixel-wise object segmentation and optical flow in videos, which is pre-trained offline to learn a generic notion, and fine-tuned online for specific objects. Caelles \etal \cite{DBLP:conf/cvpr/CaellesMPLCG17} design the one-shot video object segmentation (OSVOS) approach based on a fully-convolutional neural network to transfer generic semantic information to tackle the video object segmentation task. After that, Voigtlaender \etal \cite{DBLP:journals/corr/VoigtlaenderL17} improve the OSVOS method by updating the network online using training examples selected based on the confidence of the network and the spatial configuration. The online updating strategy noticeably improves the accuracy but sacrifices the running efficiency. To tackle time-consuming finetuning stage in the first frame, Cheng \etal \cite{DBLP:conf/cvpr/Cheng18} propose a fast VOS approach, which is formed by three modules, \ie, the part-based tracking, region-of-interest segmentation, and similarity-based aggregation. This method is able to immediately start to segment a specific object through the entire video fast and accurately. In \cite{DBLP:conf/nips/HuHS17}, a recurrent neural net approach is proposed to fuse the outputs of a binary segmentation net providing a mask and a localization net providing a bounding box for each object instance in each frame, which is able to take advantage of long-term temporal structures of the video data as well as rejecting outliers.
Bao \etal \cite{DBLP:conf/cvpr/abs-1803-09453} propose a spatio-temporal Markov Random Field (MRF) model for VOS, which uses a CNN to encode the spatial dependencies among pixels, and optical flow to establish the temporal dependencies. An efficient CNN-embedded algorithm is presented to perform approximate inference in the MRF to complete the VOS task.

{\noindent \textbf{Unsupervised video segmentation.}}
Some unsupervised video segmentation algorithms use the bottom-up strategy to group spatial-temporal coherent tubes without any prior information. Xu \etal \cite{DBLP:conf/eccv/XuXC12} implement a graph-based hierarchical segmentation method within the streaming framework, which enforces a Markovian assumption on the video stream to approximate full video segmentation. Yu \etal \cite{DBLP:conf/iccv/YuLZS15} propose an efficient and robust video segment algorithm based on parametric graph partitioning, that identifies and removes between-cluster edges to generate node clusters to complete video segmentation.

Several other unsupervised video segmentation methods upgrade bottom-up video segmentation to object-level segments. Lee \etal \cite{DBLP:conf/iccv/LeeKG11} use the static and dynamic cues to identify object-like regions in any frame, and discover hypothesis object groups with persistent appearance and motion. Then, each ranked hypothesis is used to estimate a pixel-level object labeling across all frames. Li \etal \cite{DBLP:conf/iccv/LiKHTR13} track multiple holistic figure-ground segments simultaneously to generate video object proposals, which trains an online non-local appearance models for each track using a multi-output regularized least squares formulation. Papazoglou \etal \cite{DBLP:conf/iccv/Anestis13} present a fast unsupervised VOS method, which simply aggregates the pixels in video by combining two kinds of motion boundaries extracted from optical flow to generate the proposals. In \cite{DBLP:conf/cvpr/XiaoFY16}, a series of easy-to-group instances of an object are discovered, and the appearance model of the instances are iteratively updated to detect harder instances in temporally-adjacent frames. Tokmakov \etal \cite{DBLP:conf/cvpr/TokmakovAS17} use a fully convolutional network to learn motion patterns in videos to handle VOS, which designs an encoder-decoder style architecture to first learn a coarse representation of the optical flow field features, and then refine it iteratively to produce motion labels at high-resolution.

\section{Spatiotemporal CNN for VOS}
As described above, we design a spatiotemporal CNN for VOS. Specifically, given a video sequence ${\cal X}=\{ X_1, \cdots, X_i, \cdots \}$, we aim to use our STCNN model to generate the segmentation results, \ie, ${\cal S}=\{S_1, \cdots, S_i, \cdots \}$, where $S_i$ is the segmentation mask corresponding to $X_i$. At time $t$, STCNN takes the previous $\delta$ frames, \ie, $X_{t-\delta}, \cdots, X_{t-1}$, and the current frame $X_t$, to predict the segmentation results at current frame $S_t$\footnote{For the time index $t < \delta$, we copy the first frame $\delta-t$ times to get the $\delta$ frames for segmentation.}. As shown in Figure \ref{fig:architecture}, STCNN is formed by two branches, \ie, the temporal coherence branch and the spatial segmentation branch. The temporal coherence branch learns the spatiotemporal discriminative features to capture the dynamic appearance and motion cues of video sequences instead of using optical flow. Meanwhile, the spatial segmentation branch is a fully convolutional network designed to segment objects with temporal constraints from the temporal coherence branch. In the following sections, we will describe these two branches in detail.

\subsection{Temporal Coherence Branch}
{\flushleft \textbf{Architecture.}}
As shown in Figure \ref{fig:architecture}, we construct the temporal coherence branch based on the backbone ResNet-101 network \cite{DBLP:conf/cvpr/HeZRS16}, with the input number of channels $3\delta$. That is, we concatenate the previous $\delta$ frames and feed them into the temporal coherence branch for prediction. After that, we use three deconvolution layers with the kernel size $3\times3$. To preserve spatiotemporal information in each resolution, we use three skip connections to concatenate low layer features. The convolution layer with kernel size $1\times1$ is used to compact features for efficiency. Notably, each convolution or deconvolution layer is followed by a batch normalization layer \cite{DBLP:conf/icml/IoffeS15} and a ReLU layer for non-linearity.

{\flushleft \textbf{Pretraining.}}
Motivated by \cite{DBLP:conf/iccv/JinLXSLYCDLJFY17}, we use the adversarial manner to train the temporal coherence branch by predicting future frames from unlabeled video data. Specifically, we set the temporal coherence branch as the generator ${\cal G}$, and construct a discriminator ${\cal D}$ to identify the generated video frames from ${\cal G}$ and the real video frames. Here, we use the Inception-v3 network \cite{DBLP:conf/cvpr/SzegedyVISW16} pretrained on the ILSVRC CLS-LOC dataset \cite{RussakovskyDSKS15}. We replace the last fully connected (FC) layer by a randomly initialized $2$-class FC layer as the discriminator ${\cal D}$.

At time $t$, we use the generator ${\cal G}$ to produce the prediction $\hat{X}_t$ of the current frame, based on previous $\delta$ frames $X_{t-\delta},\cdots,X_{t-1}$, \ie, $\hat{X}_{t}={\cal G}(\{X_{t-i}\}_{i=1}^{\delta})$. Then, the discriminator ${\cal D}$ is adopted to distinguish the generated frame $\hat{X}_{t}$ from the real one $X_{t}$. The generator ${\cal G}$ and discriminator ${\cal D}$ are trained iteratively in an adversarial manner \cite{DBLP:conf/nips/GoodfellowPMXWOCB14}. That is, for the fixed parameter ${\it W}_{{\cal G}}$ of the generator ${\cal G}$, we aims to optimize the discriminator ${\cal D}$ to minimize the probability of making mistakes, which is formulated as:
\begin{equation}
\begin{array}{cl}
\min_{{\it W}_{{\cal D}}}-\log\big(1-{\cal D}(\hat{X}_{t})\big) - \log{\cal D}(X_{t})
\end{array}
\end{equation}
where $\hat{X}_{t}={\cal G}(\{X_{t-i}\}_{i=1}^{\delta})$ is the generated frame from ${\cal G}$ based on previous $\delta$ frames, and $X_{t}$ is the real video frame.
Meanwhile, for the fixed parameter ${\it W}_{{\cal D}}$ of the discriminator ${\cal D}$, we expect the generator ${\cal G}$ to generate a video frame more like a real one, \ie,
\begin{equation}
\begin{array}{cl}
\min_{{\it W}_{\cal G}} \| {\it X}_{t} - \hat{X}_{t} \|_2 - \lambda_{\text{adv}}\cdot\log{\cal D}(\hat{X}_{t})
\end{array}
\end{equation}
where the first term is the mean square error, penalizing the differences between the fake frame $\hat{X}_{t}$ and the real frame $X_{t}$, the second term is the adversarial term used to maximize the probability of ${\cal D}$ making a mistake, and $\lambda_{\text{adv}}$ is the predefined parameter used to balance these two terms. In this way, the discriminator ${\cal D}$ and generator ${\cal G}$ are optimized iteratively to make the generator ${\cal G}$ capturing the discriminative spatiotemporal features in the video sequences.

\subsection{Spatial Segmentation Branch}
The spatial segmentation branch is constructed based on the ResNet-101 network \cite{DBLP:conf/cvpr/HeZRS16} by replacing the convolution layers in the last two residual blocks (\ie, {\tt res4} and {\tt res5}) with the dilated convolution layers \cite{DBLP:journals/corr/ChenPKMY14} of stride $1$, which aims to preserve the high resolution for segmentation accuracy. Then, we use the PPM module \cite{DBLP:conf/cvpr/ZhaoSQWJ17} to exploit the global context information by different-region-based context aggregation, followed by three designed attention modules to refine the predictions. That is, we apply the attention modules sequentially on multi-scale feature maps to help the network focus on object regions and ignore the background regions. After that, we concatenate the multi-scale feature maps, followed by a $3\times3$ convolution layer to produce the final prediction, see Figure \ref{fig:architecture}.

\begin{figure}
\centering
\includegraphics[width=0.85\linewidth]{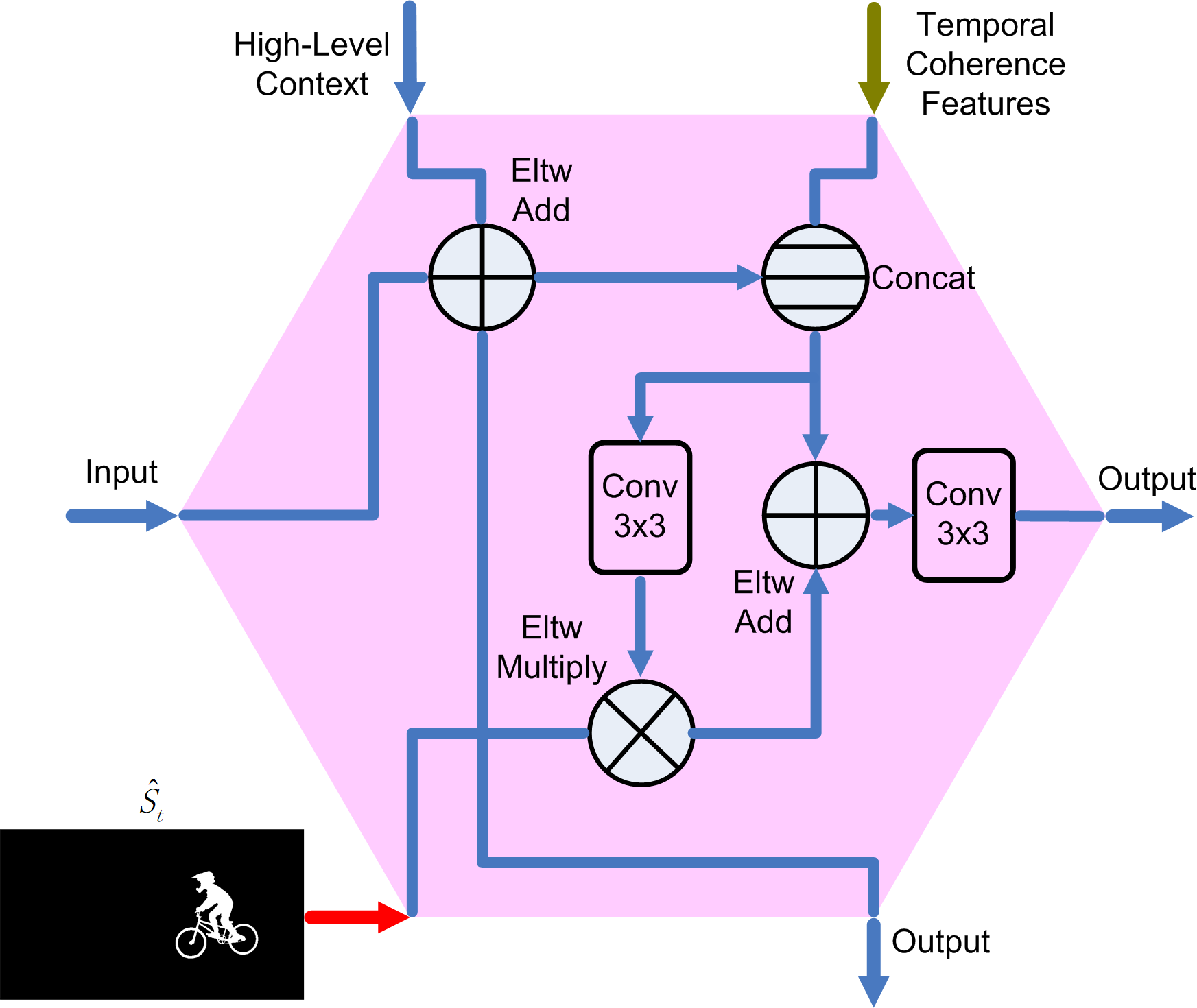}
\vspace{-2mm}
\caption{The architecture of the attention module. $\hat{{\it S}}_{t}$ denotes the segmented mask in the current stage.}
\label{fig:attention}
\end{figure}

Notably, we design the attention module to focus on object regions for accurate results. As shown in Figure \ref{fig:attention}, we first use the element-wise addition to exploit high-level context, and concatenate the temporal coherence features to integrate temporal constraints. After that, we use the predicted mask from the previous coarse scale feature map to guide the attention of the network, \ie, use the element-wise multiplication to mask the feature map in the current stage. Let $\hat{{\it S}}_{t}$ to be the predicted mask at current stage. We multiply $\hat{{\it S}}_{t}$ on the feature map in element-wise and add it to the concatenated features for prediction. In this way, the features around the object regions are enhanced, which enforces the network gradually to concentrate on object regions for accurate results.

The pixel-wise binary cross-entropy with the softmax function ${\it P}(\cdot)$ is used in multi-scale feature maps to guide the network training, see Figure \ref{fig:architecture}, which is defined as,
\begin{equation}
\begin{array}{cl}
{\cal L}(S_t, S^\ast_t) &= - \sum_{{\ell}^\ast_{i,j,t}=1} \log {\it P}(\ell_{i,j,t}=1) \\
&- \sum_{\ell^\ast_{i,j,t}=0} \log {\it P}(\ell_{i,j,t}=0)
\end{array}
\end{equation}
where $\ell^\ast_{i,j,t}$ and $\ell_{i,j,t}$ are the labels of the ground-truth mask $S^\ast_t$ and the predicted mask $S_t$ at the coordinate $(i,j)$, $\ell_{i,j,t}=1$ indicates that the prediction is foreground at the coordinate $(i,j)$, and $\ell_{i,j,t}=0$ indicates that the prediction is background at the coordinate $(i,j)$.

\begin{table*}[!t]
\caption{Performance on the validation set of DAVIS-2016. The performance of the semi-supervised VOS methods are shown in the left part, while the performance of the unsupervised VOS methods are shown in the right part. The symbol $\uparrow$ means higher scores indicate better performance, while $\downarrow$ means lower scores indicate better performance. In the last row, the numbers in parentheses are running time reported in the original papers of the corresponding methods.}
\vspace{-2mm}
\setlength{\tabcolsep}{3.0pt}
\footnotesize{
\begin{tabular}{cc|c|ccccccc|cccc}
\hline
\multicolumn{2}{c|}{\multirow{2}{*}{Metric}} &\multicolumn{8}{c}{Semi-supervised} &\multicolumn{4}{c}{Unsupervised} \\
\cline{3-14}
& &Ours &CRN\cite{DBLP:conf/cvpr/HuWKKT18} &OnAVOS\cite{DBLP:journals/corr/VoigtlaenderL17} &OSVOS\cite{DBLP:conf/cvpr/CaellesMPLCG17} &MSK\cite{DBLP:conf/cvpr/PerazziKBSS17} &CTN\cite{DBLP:conf/cvpr/JangK17} &SegFlow\cite{DBLP:conf/iccv/ChengTW017} &VPN \cite{DBLP:conf/cvpr/JampaniGG17} &ARP\cite{DBLP:conf/cvpr/KohK17} &LVO\cite{DBLP:conf/iccv/TokmakovAS17} &FSEG\cite{DBLP:conf/cvpr/JainXG17} &LMP\cite{DBLP:conf/cvpr/TokmakovAS17} \\
\hline
\hline
${\cal J}$ &Mean ($\uparrow$) &0.838 &0.844 &{\bf 0.861} &0.798 &0.797 &0.735 &0.761 &0.750 &0.762 &0.759 &0.707 &0.700 \\
&Recall ($\uparrow$) &0.961 &{\bf 0.971} &0.961 &0.936 &0.931 &0.874 &0.906 &0.901 &0.911 &0.891 &0.835 &0.850 \\
&Decay ($\downarrow$) &{\bf 0.049} &0.056 &0.052 &0.149 &0.089 &0.156 &0.121 &0.093 &0.007 &0.000 &0.015 &0.013 \\
\hline
${\cal F}$ &Mean ($\uparrow$) &0.838 &{\bf 0.857} &0.849 &0.806 &0.754 &0.693 &0.760 &0.724 &0.706 &0.721 &0.653 &0.659 \\
&Recall ($\uparrow$) &0.915 &{\bf 0.952} &0.897 &0.926 &0.871 &0.796 &0.855 &0.842 &0.835 &0.834 &0.738 &0.792 \\
&Decay ($\downarrow$) &0.064 &{\bf 0.052} &0.058 &0.150 &0.090 &0.129 &0.104 &0.136 &0.079 &0.013 &0.018 &0.025 \\
\hline
\multicolumn{2}{c|}{${\cal T}$ ($\downarrow$)} &0.191 &- &0.190 &0.376 &0.189 &0.198 &{\bf 0.182} &0.300 &0.359 &0.255 &0.295 &0.688 \\
\hline
\multicolumn{2}{c|}{Time(s/f)} &3.90 &(0.73) &(15.57) &(9.24) &(12.0) &(1.3) &(7.9) &- &- &- &- &- \\
\hline
\end{tabular}}
\label{tab:davis2016-results}
\end{table*}

\subsection{Network Implementation and Training}
\label{sec:implementation-training}
We implement our STCNN algorithm in Pytorch \cite{paszke2017automatic}. All the training and testing codes and the trained models are available at \url{https://github.com/longyin880815/STCNN}. In training phase, we first pretrain the temporal coherence branch and the spatial segmentation branch individually, and iteratively update the models of both branches. After that, we finetune both models on each sequence for online processing.

{\noindent \textbf{Pretraining temporal coherence branch.}}
We pretrain the temporal coherence branch in the adversarial manner on the training and validation sets of the ILSVRC 2015 VID dataset \cite{RussakovskyDSKS15}, which consists of $4,417$ video clips in total, \ie, $3,862$ video clips in the training set and $555$ video clips in the validation set. The backbone ResNet-101 network in our generator ${\cal G}$ is initialized by the pretrained model on the ILSVRC CLS-LOC dataset \cite{RussakovskyDSKS15}, and the other convolution and deconvolution layers are randomly initialized by the method \cite{DBLP:conf/iccv/HeZRS15}. While the discriminator ${\cal D}$ is initialized by the pretrained model on the ILSVRC CLS-LOC dataset \cite{RussakovskyDSKS15}, with the last $2$-class FC layer initialized by the method \cite{DBLP:conf/iccv/HeZRS15}. Meanwhile, we randomly flip all frames in a video clip horizontal to augment the training data, and resize all frames to the size $(480,854)$ for training. The batch size is set to $3$, and the Adam optimization algorithm \cite{DBLP:journals/corr/KingmaB14} is used to train the model. We set $\delta$ to $4$, and use the learning rates $10^{-7}$ and $10^{-4}$ to train the generator ${\cal G}$ and the discriminator ${\cal D}$, respectively. The adversarial weight $\lambda_{\text{adv}}$ is set to $0.001$ in training phase.

{\noindent \textbf{Pretraining spatial segmentation branch.}}
We use the MSRA10K salient object dataset \cite{DBLP:journals/pami/ChengMHTH15} and the PASCAL VOC 2012 segmentation dataset \cite{DBLP:journals/ijcv/EveringhamEGWWZ15} to pretrain the spatial segmentation branch. The MSRA10K dataset contains $10,000$ images, and the PASCAL VOC 2012 dataset contains $11,355$ images. Meanwhile, we randomly flip the images horizontally, and rotate the images to augment the data for training. Each training image is resized to $(300,300)$. The SGD algorithm with the batch size $8$ and learning rate $10^{-3}$ is used to optimize the model. In addition, we directly add the cross-entropy losses on multi-scale predictions (see Figure \ref{fig:architecture}) to compute the overall loss for training.

{\noindent \textbf{Iterative offline training for VOS.}}
After pretraining, we jointly finetune the model on the training set of DAVIS-2016 \cite{DBLP:conf/cvpr/PerazziPMGGS16} for VOS, which includes $30$ video clips. Specifically, we train the temporal coherence branch and the spatial segmentation branch iteratively. When optimizing the temporal coherence branch, we freeze the weights of the spatial segmentation branch, and use the learning rates $10^{-8}$ and $10^{-4}$ to train the generator ${\cal G}$ and the discriminator ${\cal D}$, respectively. The Adam algorithm is used to optimize the weights in temporal coherence branch with the batch size $1$. For training the spatial segmentation branch, similarly we fix the weights in the temporal coherence branch and only update the weights in the spatial segmentation branch using the SGD algorithm with the learning rate $10^{-4}$. For better training, we randomly flip horizontally, rotate and rescale to augment the training data. For this iterative learning process, each branch in the network is able to obtain useful information from another branch through back-propagation. In this way, the spatial segmentation branch can receive useful temporal information from the temporal coherence branch, while the temporal coherence branch can learn more effective spatiotemporal features for accurate segmentation.

{\noindent \textbf{Online training for VOS.}}
To adapt the network to a specific object for VOS, we finetune the network on the first frame for each video clip. Since we only have the annotation mask in the first frame, only the spatial segmentation branch is optimized. Each mask in the first frame is augmented to generate multiple training samples to increase the diversity. Specifically, we use the ``lucid dream'' strategy \cite{Khoreva17davis} to generate in-domain training data based on the provided annotation in the first frame, including $5$ steps, \ie, illumination changing, foreground-background splitting, object motion simulating, camera view changing, and foreground-background merging. Notably, in contrast to \cite{Khoreva17davis}, we do not generate the optical flow since our STCNN do not require the optical flow for video segmentation. The SGD algorithm with the learning rate $10^{-4}$ and batch size $1$ is used to train the network online.

\begin{table*}
\caption{The results on the Youtube-Objects dataset. The mean intersection-over-union is used to evaluate the performance of methods. The results are directly taken from the original paper. The symbol $\uparrow$ means higher scores indicate better performance. Bold font indicates the best result.}
\vspace{-2mm}
\setlength{\tabcolsep}{3.5pt}
\footnotesize{
\begin{tabular}{c|ccccccccccc|c}
\hline
Method              &BVS\cite{DBLP:conf/cvpr/MarkiPWS16} &JFS\cite{DBLP:conf/iccv/NagarajaSB15} &SCF\cite{DBLP:conf/eccv/JainG14} &MRFCNN\cite{DBLP:conf/cvpr/abs-1803-09453} &LT\cite{Khoreva17davis} &OSVOS\cite{DBLP:conf/cvpr/CaellesMPLCG17} &MSK\cite{DBLP:conf/cvpr/PerazziKBSS17} &OFL\cite{DBLP:conf/cvpr/Tsai0B16} &CRN\cite{DBLP:conf/cvpr/HuWKKT18} &DRL\cite{DBLP:conf/cvpr/HanLZX18} &OnAVOS\cite{DBLP:journals/corr/VoigtlaenderL17} &Ours \\
\hline
\hline
aeroplane &0.868 &0.890 &0.863 &- &- &0.868 &0.845 &{\bf 0.899} &- &0.852 &- &0.869 \\
bird       &0.809 &0.816 &0.810 &- &- &0.851 &0.837 &0.842 &- &0.868 &- &{\bf 0.879} \\
boat      &0.651 &0.742 &0.686 &- &- &0.754 &0.774 &0.740 &- &{\bf 0.799} &- &0.786 \\
car        &0.687 &0.709 &0.694 &- &- &0.709 &0.640 &0.809 &- &0.672 &- &{\bf 0.859} \\
cat        &0.559 &0.677 &0.589 &- &- &0.676 &0.698 &0.683 &- &0.746 &- &{\bf 0.772} \\
cow       &0.699 &0.791 &0.686 &- &- &0.762 &0.767 &{\bf 0.798} &- &0.746 &- &0.781 \\
dog       &0.685 &0.703 &0.618 &- &- &0.779 &0.745 &0.766 &- &{\bf 0.827} &- &0.800 \\
horse     &0.589 &0.678 &0.540 &- &- &0.714 &0.641 &0.726 &- &0.736 &- &{\bf 0.738} \\
motorbike &0.605 &0.615 &0.609 &- &- &0.582 &0.892 &0.481 &- &{\bf 0.737} &- &0.680 \\
train       &0.652 &0.782 &0.663 &- &- &0.746 &0.744 &0.763 &- &{\bf 0.830} &- &0.796 \\
\hline
Mean ($\uparrow$)   &0.680 &0.740 &0.676 &0.784 &0.762 &0.744 &0.717 &0.776 &0.766 &0.781 &0.774 &{\bf 0.796} \\
\hline
\end{tabular}}
\label{tab:youtube-results}
\end{table*}

\section{Experiment}
We evaluate the proposed algorithm against state-of-the-art VOS methods on three challenging datasets, namely the DAVIS-2016 \cite{DBLP:conf/cvpr/PerazziPMGGS16}, DAVIS-2017 \cite{DBLP:journals/corr/Pont-TusetPCASG17}, and Youtube-Object \cite{DBLP:conf/cvpr/PrestLCSF12,DBLP:conf/eccv/JainG14}. All the experiments are conducted on a workstation with a 3.6 GHz Intel i7-4790 CPU, 16GB RAM, and a NVIDIA Titan 1080ti GPU. The quantitative results are presented in Table \ref{tab:davis2016-results} and \ref{tab:youtube-results}. Some qualitative segmentation results are shown in Figure \ref{fig:qualitative-results}, and more video segmentation results can be found in supplementary material.

\subsection{DAVIS-2016 Dataset}
The DAVIS-2016 dataset \cite{DBLP:conf/cvpr/PerazziPMGGS16} comprises of $50$ sequences, $3,455$ annotated frames with a binary pixel-level foreground/background mask. Due to the computational complexity being a major bottleneck in video processing, the sequences in the dataset have a short temporal extent (about $2$-$4$ seconds), but include all major challenges typically found in longer video sequences, such as background clutter, fast-motion, edge ambiguity, camera-shake, and out-of-view. We tested the proposed method on the 480p resolution set.

\subsubsection{Evaluation}
For comprehensive evaluation, we use three measures provided by the dataset, \ie, region similarity ${\cal J}$, contour accuracy ${\cal F}$ and temporal instability ${\cal T}$. Specifically, region similarity ${\cal J}$ measures the number of mislabeled pixels, which is defined as the intersection-over-union (IoU) of the estimated segmentation and the ground-truth mask. Given a segmentation mask ${\it S}$ and the ground-truth mask ${\it S}^\ast$, ${\cal J}$ is calculated as ${\cal J}=\frac{{\it S}\cap{\it S}^\ast}{{\it S}\cup{\it S}^\ast}$. The contour accuracy ${\cal F}$ computes the F-measure of the contour-based precision ${\it P}_c$ and recall ${\it R}_c$ between the contour points of estimated segmentation ${\it S}$ and the ground-truth mask ${\it S}^\ast$, defined as ${\cal F}=\frac{2{\it P}_c{\it R}_c}{{\it P}_c+{\it R}_c}$. In addition, the temporal instability ${\cal T}$ measures oscillations and inaccuracies of the contours, which is calculated by following \cite{DBLP:conf/cvpr/PerazziPMGGS16}.

\subsubsection{Ablation Study}
To comprehensively understand the proposed method, we conduct several ablation experiments. Specifically, we construct three variants and evaluate them on the {\tt validation} set of DAVIS-2016, to validate the effectiveness of different components (\ie, the ``Lucid dream'' augmentation, the attention module, and the temporal coherence branch) in the proposed method, shown in Table \ref{tab:module-ablation}. Meanwhile, we also conduct experiments to analyze the importance of different training phases in Table \ref{tab:training-ablation}. For a fair comparison, we use the same parameter settings except for the specific declaration.

{\noindent \textbf{Lucid dream augmentation.}}
To demonstrate the effect of the ``Lucid Dream'' augmentation, we remove it from our STCNN model (see the forth column in Table \ref{tab:module-ablation}). As shown in Table \ref{tab:module-ablation}, we find that the region similarity ${\cal J}$ is reduced from $0.838$ to $0.832$. This decline (\ie, $0.006$) demonstrate that the ``Lucid dream'' data augmentation is useful to improve the performance.

{\noindent \textbf{Attention module.}}
To validate the effectiveness of the attention module, we construct an algorithm by further removing the attention mechanism in the spatial segmentation branch. That is, we remove the red lines in Figure \ref{fig:architecture} to directly generate the output mask. In this way, the object region is not specifically concentrated by the network. The segmentation results of the model is reported in the third column in Table \ref{tab:module-ablation}. We compare the third and forth columns in Table \ref{tab:module-ablation}, and find that the attention module improves $0.01$ region similarity ${\cal J}$, and $0.015$ contour accuracy ${\cal F}$, which demonstrates that the attention module is critical to the performance. The main reason is that the attention module is gradually applied on multi-scale features maps, enforcing the network to focus on the object regions to generate more accurate results.

\begin{figure*}[t]
\centering
\includegraphics[width=1.0\linewidth]{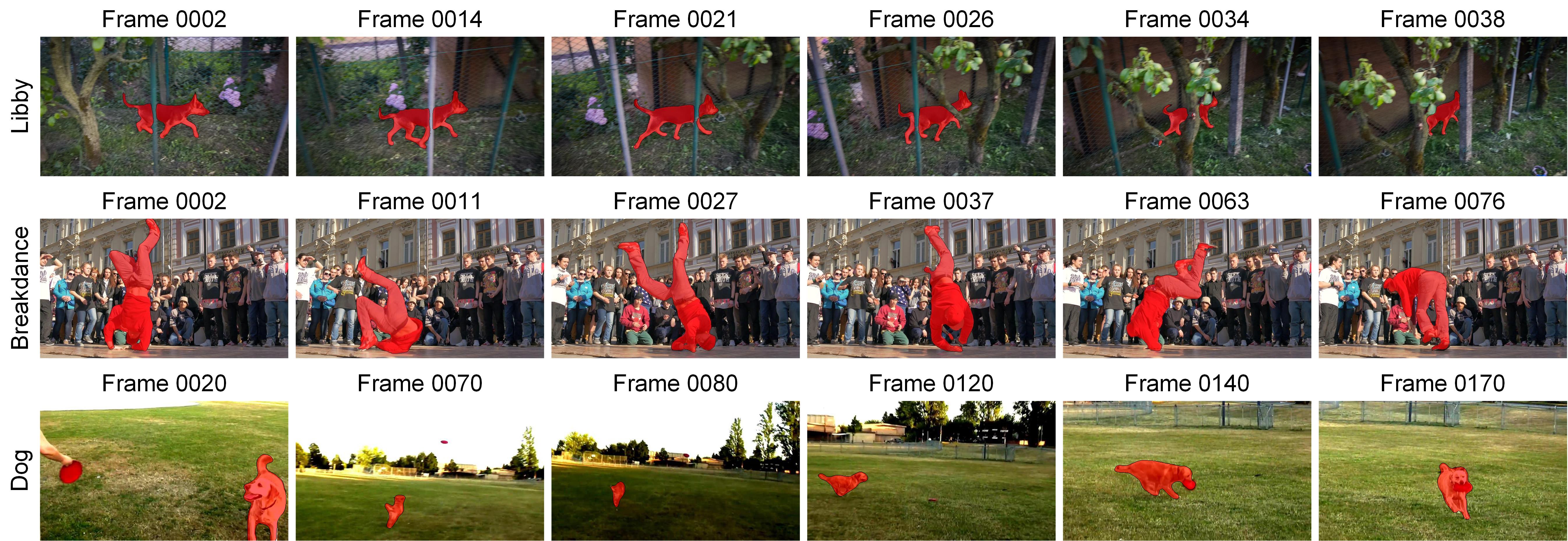}
\caption{The qualitative segmentation results of STCNN on the DAVIS-2016 (first two rows) and Youtube-Objects (last row) datasets. The output on the pixel level are indicated by the red mask. The results show that our method is able to segment objects under several challenges, such as occlusions, deformed shapes, fast motion, and cluttered backgrounds.}
\label{fig:qualitative-results}
\end{figure*}

{\noindent \textbf{Temporal coherence branch.}}
We construct a network based on the spatial segmentation branch without the attention module and report its results in the second column in Table \ref{tab:module-ablation}. Comparing the results between the second and third columns in Table \ref{tab:module-ablation}, we observe that the temporal coherence branch is critical to the performance of video segmentation, \ie, it improves $0.01$ mean region similarity ${\cal J}$ ($0.812$ {\em vs.} $0.822$) and $0.013$ mean contour accuracy ${\cal F}$ ($0.807$ {\em vs.} $0.820$). Most importantly, the temporal coherence branch significantly reduces the temporal instability, \ie, it reduces relative $13.4\%$ temporal instability ${\cal T}$ ($0.231$ {\em vs.} $0.200$). The results demonstrate that the temporal coherence branch is effective to capture the dynamic appearance and motion cues of video sequences to help generate accurate and consistent segmentation results.

{\flushleft \textbf{Training analysis.}}
As described in Section \ref{sec:implementation-training}, we first iteratively update the pretrained temporal coherence branch and the spatial segmentation branch offline. After that, we finetune both branches on each sequence for online processing. We evaluate the proposed STCNN method with different training phase on the {\tt validation} set of DAVIS-2016 to analyze their effects on performance in Table \ref{tab:training-ablation}. As shown in Table \ref{tab:training-ablation}, we find that without online training phase, the mean region similarity ${\cal J}$ of STCNN drops $0.096$ (\ie, $0.838$ {\em vs.} $0.742$), while without offline training phase, ${\cal J}$ of STCNN drops $0.052$ (\ie, $0.838$ {\em vs.} $0.786$). In summary, both training phases are extremely important to our STCNN, especially for the online training phase.

\begin{table}[t]
\centering
\caption{Effectiveness of various components in the proposed method. All models are evaluated on the DAVIS-2016 dataset. The symbol $\uparrow$ means high scores indicate better result, while $\downarrow$ means lower scores indicate better result.}
\vspace{-2mm}
\footnotesize \setlength{\tabcolsep}{7.0pt}
\begin{tabular}{c|cccc}
\hline
\multicolumn{1}{c|}{Component} &\multicolumn{4}{c}{STCNN} \\
\hline
Temporal Coherence Branch?  &  &\Checkmark &\Checkmark &\Checkmark  \\
Attention Module? &  & & \Checkmark & \Checkmark \\
Lucid Dream?   & & & & \Checkmark\\
\hline
 ${\cal J}$ Mean ($\uparrow$) &0.812  &0.822  &0.832  &0.838 \\
 ${\cal F}$ Mean ($\uparrow$) &0.807  &0.820  &0.835  &0.838 \\
 ${\cal T}$ ($\downarrow$)      &0.231  &0.200  &0.192  &0.191  \\
\hline
\end{tabular}
\label{tab:module-ablation}
\end{table}

\subsubsection{Comparison with State-of-the-Arts}
We compare the proposed method with $7$ state-of-the-art semi-supervised methods, \ie, CRN \cite{DBLP:conf/cvpr/HuWKKT18}, OnAVOS \cite{DBLP:journals/corr/VoigtlaenderL17}, OSVOS \cite{DBLP:conf/cvpr/CaellesMPLCG17}, MSK \cite{DBLP:conf/cvpr/PerazziKBSS17}, CTN \cite{DBLP:conf/cvpr/JangK17}, SegFlow \cite{DBLP:conf/iccv/ChengTW017}, and VPN \cite{DBLP:conf/cvpr/JampaniGG17}, and $4$ state-of-the-art unsupervised methods, namly ARP \cite{DBLP:conf/cvpr/KohK17}, LVO \cite{DBLP:conf/iccv/TokmakovAS17}, FSEG \cite{DBLP:conf/cvpr/JainXG17}, and LMP \cite{DBLP:conf/cvpr/TokmakovAS17} in Table \ref{tab:davis2016-results}.

As shown in Table \ref{tab:davis2016-results}, our algorithm outperforms the existing semi-supervised algorithms (\eg, OSVOS \cite{DBLP:conf/cvpr/CaellesMPLCG17} and MSK \cite{DBLP:conf/cvpr/PerazziKBSS17}) and unsupervised algorithms (\eg, ARP \cite{DBLP:conf/cvpr/KohK17} and LVO \cite{DBLP:conf/iccv/TokmakovAS17}) with $0.838$ mean region similarity ${\cal J}$, $0.838$ mean contour accuracy ${\cal F}$, and $0.191$ temporal stability ${\cal T}$, except CRN \cite{DBLP:conf/cvpr/HuWKKT18} and OnAVOS \cite{DBLP:journals/corr/VoigtlaenderL17}. The OnAVOS algorithm \cite{DBLP:journals/corr/VoigtlaenderL17} updates the network online using training examples selected based on the confidence of the network and the spatial configuration, which requires heavy consumption of time and computation resource. Our algorithm is much more efficient and do not require the optical flow in both training and testing phase. The online updating mechanism in OnAVOS \cite{DBLP:journals/corr/VoigtlaenderL17} is complementary to our method. We believe that is can be used in our STCNN to further improve the performance. In addition, in contrast to CRN \cite{DBLP:conf/cvpr/HuWKKT18} relying on optical flow to render temporal coherence in both training and testing, our method uses a self-supervised strategy to implicitly exploit the temporal coherence without relying on the expensive human annotations of optical flow. The temporal coherence branch is able to capture the dynamic appearance and motion cues of video sequences, pretrained in an adversarial manner from nearly unlimited unlabeled video data.

\subsubsection{Runtime Performance}
We present the inference time of STCNN and the state-of-the-art methods on the {\tt validation} set of DAVIS-2016 in the last row of Table \ref{tab:davis2016-results}. Since different algorithms are developed and evaluated on different platforms (\eg, different algorithms are evaluated on different types of GPUs), it is difficult to compare the running time efficiency fairly. We report the running speed for reference. Meanwhile, we also analyze the influence of the number of iterations in the online training phase of STCNN to the segmentation accuracy and running speed in Table \ref{tab:iteration-accuracy}. With the number of iterations increasing, the mean region similarity ${\cal J}$ increases to reach a maximal value $0.838$. Continue training is not able to obtain the accuracy gain, but slows down the inference speed. Thus, we set the number of iterations in online training to $400$ in our experiments. Compared to the state-of-the-art methods such as OSVOS ($9.24$ s/f), OnAVOS ($15.57$ s/f), our method achieves impressive results with much faster running speed.

\begin{table}[t]
\centering
\caption{Performance and running speed of the proposed STCNN with different number of iterations in online training phase.}
\vspace{-2mm}
\footnotesize \setlength{\tabcolsep}{7.0pt}
\begin{tabular}{c|cccccc}
\hline
${\cal \#}$Iter &100 &200 &300 &400 &500 &600 \\
\hline
\hline
mIOU  &0.830 &0.834 &0.836 &0.838 &0.838 &0.838  \\
\hline
time(s/f) &1.11  &2.04 &2.97  &3.90 &4.83 &5.76 \\
\hline
\end{tabular}
\label{tab:iteration-accuracy}
\end{table}

\begin{table}[t]
\centering
\caption{Performance on DAVIS-2016 for different training phases of STCNN.}
\vspace{-2mm}
\footnotesize \setlength{\tabcolsep}{12.0pt}
\begin{tabular}{cc|ccc}
\hline
& &Offline &Online  & \\
&Metric& Training &Training&All\\
\hline
\hline
${\cal J}$ &Mean ($\uparrow$) &0.742 &0.786 &0.838\\
&Recall ($\uparrow$)&0.854&0.921&0.961\\
&Decay ($\downarrow$)&-0.004 &0.075 &0.049\\
\hline
${\cal F}$ &Mean ($\uparrow$)&0.743&0.79 &0.838\\
&Recall ($\uparrow$) &0.806&0.871&0.915\\
&Decay ($\downarrow$)&0.018&0.089&0.064\\
\hline
\end{tabular}
\label{tab:training-ablation}
\end{table}


\subsection{DAVIS-2017 Dataset}
We evaluate our STCNN on the DAVIS-2017 validation set \cite{DBLP:journals/corr/Pont-TusetPCASG17}, which consists of $30$ video sequences with various challenging cases including multiple objects with similar appearance, heavy occlusion, large appearance variation, clutter background, etc. The mean region similarity ${\cal J}$ and contour accuracy ${\cal F}$ are used to evaluate the performance in Table \ref{tab:davis2017-results}. Our STCNN performs favorably against most of the semi-supervised methods, \eg, OSVOS \cite{DBLP:conf/cvpr/CaellesMPLCG17}, FAVOS \cite{DBLP:conf/cvpr/ChengTHWY18}, and MSK \cite{DBLP:conf/cvpr/PerazziKBSS17}, with $58.7\%$ mean region similarity ${\cal J}$  and $64.6\%$ contour accuracy ${\cal F}$. The results demonstrate that our STCNN is effective to segment objects in more complex scenarios with similar appearance.

\begin{table}
\centering
\caption{The results on the DAVIS-2017 dataset. The symbol $\uparrow$ means higher scores indicate better performance. Bold font indicates the best result.}
\vspace{-2mm}
\footnotesize \setlength{\tabcolsep}{5.5pt}
\begin{tabular}{c|cccccc|c}
\hline
Metric &\cite{DBLP:conf/cvpr/Tsai0B16} &\cite{DBLP:conf/cvpr/PerazziKBSS17} &\cite{DBLP:conf/cvpr/YangWXYK18} &\cite{DBLP:conf/cvpr/ChengTHWY18} &\cite{DBLP:conf/cvpr/CaellesMPLCG17} &\cite{DBLP:journals/corr/VoigtlaenderL17} &Ours \\
\hline
${\cal J}$ Mean ($\uparrow$) &43.2 &51.2 &52.5 &54.6 &56.6 &{\bf 61.6} &58.7 \\
${\cal F}$ Mean ($\uparrow$) &- &57.3 &57.1 &61.8 &63.9 &{\bf 69.1} &64.6 \\
\hline
\end{tabular}
\label{tab:davis2017-results}
\end{table}

\subsection{Youtube-Objects Dataset}
The Youtube-Objects dataset \cite{DBLP:conf/cvpr/PrestLCSF12,DBLP:conf/eccv/JainG14} contains web videos from $10$ object categories. $126$ video sequences with more than $20,000$ frames and ground-truth masks provided by \cite{DBLP:conf/eccv/JainG14} are used for evaluation, where a single object or a group of objects of the same category are separated from the background. The videos in Youtube-Objects have a mix of static and moving objects, and the number of frames in each video clip ranges from $2$ to $401$. The mean IoU between the estimated results and the ground-truth masks in all video frames is used to evaluate the performance of the algorithms.

We compare the proposed STCNN method to $11$ state-of-the-art semi-supervised algorithms, namely BVS \cite{DBLP:conf/cvpr/MarkiPWS16}, JFS \cite{DBLP:conf/iccv/NagarajaSB15}, SCF \cite{DBLP:conf/eccv/JainG14}, MRFCNN \cite{DBLP:conf/cvpr/abs-1803-09453}, LT \cite{Khoreva17davis}, OSVOS \cite{DBLP:conf/cvpr/CaellesMPLCG17}, MSK \cite{DBLP:conf/cvpr/PerazziKBSS17}, OFL \cite{DBLP:conf/cvpr/Tsai0B16}, CRN \cite{DBLP:conf/cvpr/HuWKKT18}, DRL \cite{DBLP:conf/cvpr/HanLZX18}, and OnAVOS \cite{DBLP:journals/corr/VoigtlaenderL17} in Table \ref{tab:youtube-results}. As shown in Table \ref{tab:youtube-results}, we observe that the STCNN method produces the best results with $0.796$ mean IoU, which surpasses the state-of-the-art results, \ie, MRFCNN \cite{DBLP:conf/cvpr/abs-1803-09453} ($0.784$ mean IoU), with $0.012$ mIoU. Compared to the optical flow based methods \cite{DBLP:conf/cvpr/Tsai0B16,DBLP:conf/cvpr/PerazziKBSS17}, our STCNN method performs well on fast moving objects, such as {\em car} and {\em cat}. The estimation of optical flow for fast moving objets is inaccurate, affecting the segmentation accuracy. Our STCNN relies on the temporal coherence branch to capture discriminative spatiotemporal features, which is effective to tackle such scenario. Meanwhile, the algorithm \cite{DBLP:conf/eccv/JainG14} use long-term supervoxels to capture the temporal coherence. Only the superpixels are used in segmentation, causing the inaccurate boundaries of objects. In contrast, our algorithm design a coarse-to-fine process to sequentially apply the attention module on multi-scale feature maps, enforcing the network to focus on object regions to generate accurate results, especially for the non-rigid objects, \eg, {\em cat} and {\em horse}. The qualitative results are shown in the last three rows in Figure \ref{fig:qualitative-results}.


\section{Conclusion}
In this work, we present an end-to-end trained spatiotemporal CNN for VOS, which is formed by two branches, \ie, the temporal coherence branch and the spatial segmentation branch. The temporal coherence branch is pretrained in an adversarial fashion, and used to predict the appearance and motion cues in the video sequence to guide object segmentation without using optical flow. The spatial segmentation branch is designed to segment object instance accurately based on the predicted appearance and motion cues from the temporal coherence branch. In addition, to obtain accurate segmentation results, a coarse-to-fine process is iteratively applied on multi-scale feature maps in the spatial segmentation branch to refine the predictions. These two branches are jointly trained in an end-to-end manner. Extensive experimental results on three challenging datasets, \ie, DAVIS-2016, DAVIS-2017, and Youtube-Object, demonstrate that the proposed method achieves favorable performance against the state-of-the-arts.

\section*{Acknowledgments}
Kai Xu, Guorong Li, and Qingming Huang were supported by National Natural Science Foundation of China: 61772494, 61620106009, 61836002, U1636214 and 61472389, Key Research Program of Frontier Sciences, CAS: QYZDJ-SSW-SYS013, Youth Innovation Promotion Association CAS, and the University of Chinese Academy of Sciences.

{\small
\bibliographystyle{ieee}
\bibliography{reference}
}

\end{document}